\documentclass[10pt,twocolumn,letterpaper]{article}

\usepackage{iccv}
\usepackage{times}
\usepackage{epsfig}
\usepackage{graphicx}
\usepackage{amsmath}
\usepackage{amssymb}

\usepackage{booktabs} 
\usepackage{multirow} 
\usepackage{pifont} 
\usepackage{setspace} 
\usepackage{verbatim} 
\usepackage{array} 

\makeatletter
\newcommand{\myline}{
    \noalign {\ifnum 0=`}\fi \hrule height 1pt
    \futurelet \reserved@a \@xhline
}
\newcolumntype{"}{@{\hskip \tabcolsep \vrule width 1pt \hskip \tabcolsep}}
\makeatother

\usepackage[breaklinks=true,bookmarks=false]{hyperref}

\iccvfinalcopy 

\def\iccvPaperID{****} 

\ificcvfinal\fi

\begin{document}

\title{CM-NAS: Cross-Modality Neural Architecture Search \\ for Visible-Infrared Person Re-Identification}

\author{
Chaoyou Fu$^{1,2}$\thanks{Equal contribution.} \quad
Yibo Hu$^{3*}$ \quad
Xiang Wu$^{2}$ \quad
Hailin Shi$^3$ \quad
Tao Mei$^3$ \quad
Ran He$^{1,2}$\thanks{Corresponding author.} \\
$^1$ School of Artificial Intelligence, University of Chinese Academy of Sciences \\
$^2$ NLPR \& CEBSIT \& CRIPAC, CASIA ~~~~~~ $^3$ JD AI Research \\
{\tt\small 
\{chaoyou.fu, rhe\}@nlpr.ia.ac.cn, \{shihailin, tmei\}@jd.com
} \\
{\tt\small 
\{huyibo871079699,alfredxiangwu\}@gmail.com
}
}

\maketitle
\ificcvfinal\thispagestyle{empty}\fi

\begin{abstract}
Visible-Infrared person re-identification (VI-ReID) aims to match cross-modality pedestrian images, breaking through the limitation of single-modality person ReID in dark environment. In order to mitigate the impact of large modality discrepancy, existing works manually design various two-stream architectures to separately learn modality-specific and modality-sharable representations. Such a manual design routine, however, highly depends on massive experiments and empirical practice, which is time consuming and labor intensive. In this paper, we systematically study the manually designed architectures, and identify that appropriately separating Batch Normalization (BN) layers is the key to bring a great boost towards cross-modality matching. Based on this observation, the essential objective is to find the optimal separation scheme for each BN layer. To this end, we propose a novel method, named Cross-Modality Neural Architecture Search (CM-NAS). It consists of a BN-oriented search space in which the standard optimization can be fulfilled subject to the cross-modality task. Equipped with the searched architecture, our method outperforms state-of-the-art counterparts in both two benchmarks, improving the Rank-1/mAP by \textbf{6.70\%/6.13\%} on SYSU-MM01 and by \textbf{12.17\%/11.23\%} on RegDB. Code is released at~{\color{blue} \url{https://github.com/JDAI-CV/CM-NAS}}.
\end{abstract}

\section{Introduction} \label{Introduction}
Person re-identification (ReID) refers to matching pedestrian images acquired from disjoint cameras \cite{leng2019survey,yu2018unsupervised,zheng2017unlabeled}.
In recent years, it has received substantial attention due to its significant practical value in video surveillance \cite{ye2020cross}.
Conventional person ReID is only devoted to single-modality, \textit{i.e.} all the pedestrian images are taken by visible cameras during day time.
Benefiting from the strenuous efforts of researchers, impressive achievements have been made on most benchmarks \cite{zheng2016mars,bai2017scalable,zhai2020ad}.
Nevertheless, the visible cameras cannot image clearly in dark environment, which impedes the popularization and application of person ReID \cite{wang2019rgb}.
To overcome this obstacle, in addition to the visible (VIS) cameras, infrared (IR) cameras that are robust to illumination variants are also equipped in many surveillance scenarios \cite{dai2018cross}.
Therefore, in practice, we often need to match VIS and IR pedestrian images, raising the task of VI-ReID.

Modality discrepancy, caused by different wavelengths of VIS and IR images, is one of the most difficult challenges in VI-ReID.
Existing works have manually designed various two-stream architectures \cite{ye2019bi,ye2020dynamic,liu2020enhancing} to mitigate the impact of the large modality discrepancy.
Specifically, as exemplified in Fig.~\ref{fig-resnet50}, some layers are separated into two branches to learn modality-specific representations for VIS an IR data respectively, while the remaining layers are shared to learn modality-sharable representations.
The intuition behind this design is that VIS and IR images contain both modality-specific information, \textit{e.g.} the spectrum, and modality-sharable information, \textit{e.g.} the identity.
At this point, an obvious problem is raised: which layers should be separated into two branches and which layers should be shared?
Some methods separate the layers in the first one \cite{ye2020cross,ye2020dynamic} or five \cite{ye2018visible} blocks, while some others even share the whole network \cite{wu2017rgb}.
There is still no consensus on the optimal design of the neural architecture for VI-ReID.

In this paper, to investigate the impact of different separation schemes, we manually design a total of \textbf{195} different two-stream architectures.
Given that Batch Normalization (BN) plays a crucial role in learning modality distributions \cite{xie2020adversarial}, we also perform separation in units of BN layers, in addition to the entire block as usual.
As illustrated in Section~\ref{rethinking}, after comprehensively comparing the performances of all the architectures, we obtain two major observations:
(1) only separating BN layers in the block is superior than separating the entire block;
(2) separating two blocks of BN layers generally outperforms separating a single one.
With these in mind, we arrive at a conclusion that \textbf{appropriately separating all BN layers is the key to bring a great boost towards cross-modality matching}.
As a consequence, the essential objective is to find the optimal separation scheme for each BN layer in the backbone. However, there are a great deal of potential separation schemes.
Specifically, the backbone ResNet50 \cite{he2016deep} contains 53 BN layers, leading to a total of 2$^{53}$ possible architectures.
It is time consuming and labor intensive to manually traverse through all the possible architectures.
To tackle this intractable problem, inspired by recently thriving Neural Architecture Search (NAS) technique \cite{liu2018darts,hu2020tf,tan2019mnasnet,dai2019chamnet}, we propose a novel Cross-Modality NAS (CM-NAS) to automatically determine whether each BN layer separates or not.
A BN-oriented search space is elaborately built in which the standard optimization can be fulfilled subject to the cross-modality task.
Note that it is infeasible to directly apply existing single-modality NAS methods like Auto-ReID \cite{quan2019auto}, because its search space is powerless to bridge the modality discrepancy, as discussed in Section~\ref{experimental analyses}.
In contrast, our designed search space supports to learn both modality-specific and modality-sharable representations via switching corresponding separating and sharing operations, which first opens the door of NAS to cross-modality matching.

Without bells and whistles, our method exceeds all state-of-the-art methods on both two VI-ReID benchmarks.
On SYSU-MM01, our method achieves an improvement of \textbf{6.70\%} and \textbf{6.13\%} in terms of the Rank-1 accuracy and the mAP score. On RegDB, our method promotes the two indicators by \textbf{12.17\%} and \textbf{11.23\%}, respectively.
Compared with the baseline ResNet50, our method increases the Rank-1/mAP by \textbf{7.50\%/6.70\%} on SYSU-MM01 and by \textbf{8.73\%/8.35\%} on RegDB, with a small additional parameters and no extra computational costs.
We hope this simple yet effective method will be a solid foundation to facilitate future research in VI-ReID.

To sum up, we make the following three contributions:

\begin{itemize}
  \item We systematically analyze \textbf{195} different manually designed architectures, and identify the significance of separating BN layers. This conclusion motivates us to develop a BN-oriented search algorithm.

  \item A novel CM-NAS is proposed to automatically search the optimal separation scheme for BN layers, which fills the blank of NAS in cross-modality matching.

  \item Our method significantly surpasses state-of-the-art competitors in both two benchmarks, improving the Rank-1/mAP by \textbf{6.70\%/6.13\%} on SYSU-MM01 and by \textbf{12.17\%/11.23\%} on RegDB. Code will be released to aid future research in VI-ReID.
\end{itemize}

\section{Related Works}
\paragraph{Single-Modality Person ReID.}
The goal of single-modality person ReID is to match pedestrian images across non-overlapping visible cameras \cite{gong2014person,zheng2016person,zhang2019danet,gao2020interactgan}.
Existing works can be divided into three categories, including hand-crafted descriptors methods \cite{liao2015person,zheng2015scalable}, metric learning methods \cite{zheng2012reidentification,liao2015efficient} and deep learning methods \cite{sun2018beyond,zheng2019joint,wang2018mancs,he2018deep,wu2020learning}.
However, the visible cameras cannot image clearly in dark environment, which impedes the popularization and application of the single-modality person ReID \cite{zhao2020not,wang2019rgb}. 

\paragraph{Visible-Infrared Person ReID.}
Different from the aforementioned single-modality person ReID, VI-ReID matches pedestrian images belonging to different modalities.
Due to the great practical value of VI-ReID, it has received substantial attention in recent years \cite{feng2019learning,wang2019rgb,lu2020cross,ye2020dynamic}. 
The pioneer work \cite{wu2017rgb} contributes a new VI-ReID dataset named SYSU-MM01 and introduces a deep zero-padding method, which explores modality-specific information in a one-stream network.
\cite{dai2018cross} proposes a cross-modality generative adversarial framework to reduce the modality discrepancy.
\cite{ye2018visible} learns discriminative cross-modality features via an elaborately designed dual-path network and a bi-directional dual-constrained top-ranking loss.
\cite{hao2019hsme} leverages a two-stream HSMEnet to map both representation learning and metric learning on to a hypersphere manifold.

\begin{figure*}[t]
\centering
\includegraphics[width=0.97 \textwidth]{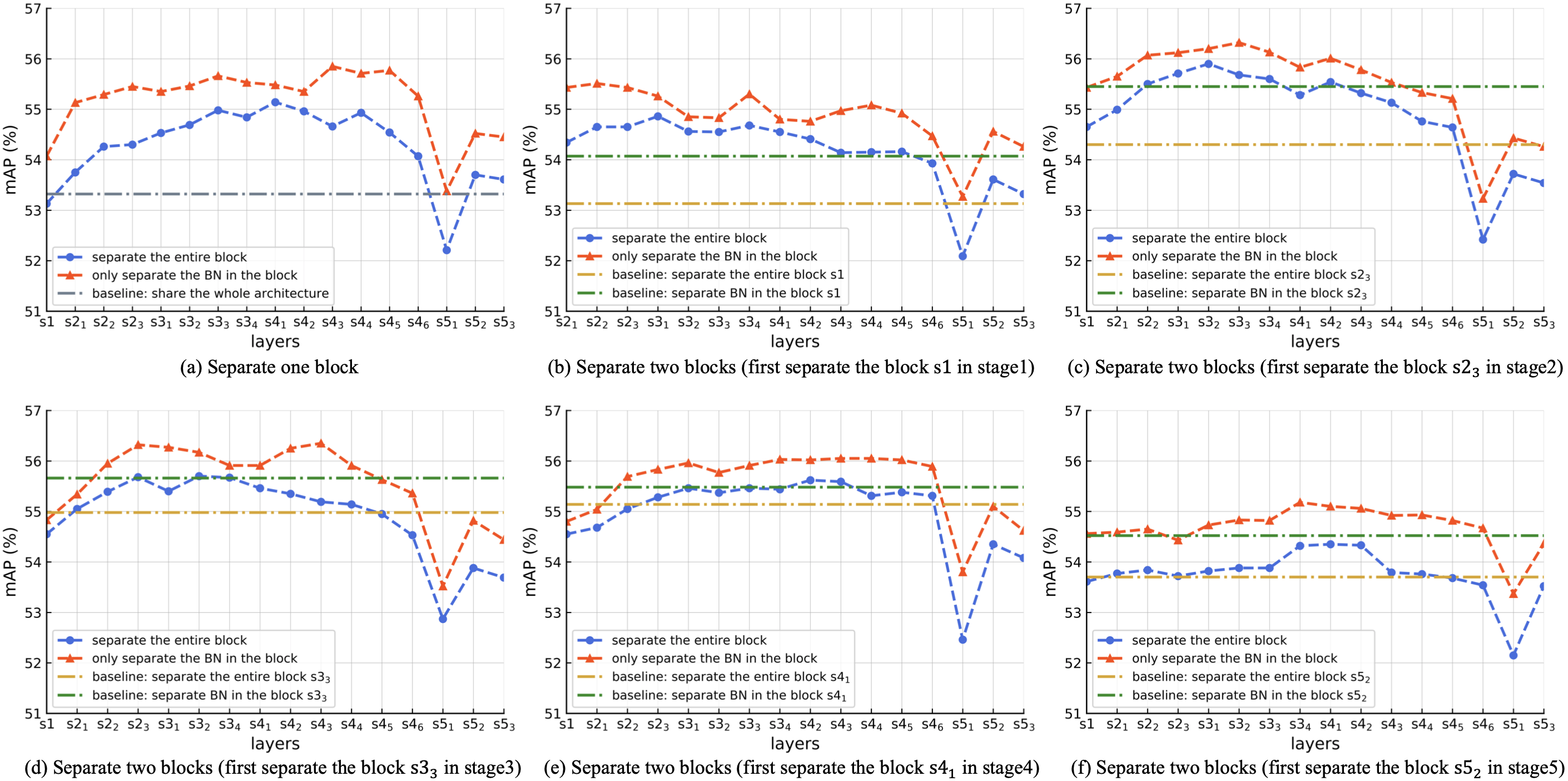}
\caption{Comparisons of different ResNet50-based architecture designs on SYSU-MM01 under the challenging \emph{single-shot}$\&$\emph{all-search} setting \cite{wu2017rgb}.
`s$2_1$' denotes the 1-st block in stage2 of ResNet50, whose architecture is depicted in Fig.~\ref{fig-resnet50}.
We design the architecture in units of the entire block (the blue line) or only the BN layers in the block (the red line).
(a) shows the results of merely separating one block. (b)-(f) present the results of separating two blocks, where we first fixedly separate a block and then traverse through the remaining stages to separate the other one. Note that for (c), we choose to first fixedly separate the block `s$2_3$' rather than `s$2_1$' or `s$2_2$' in stage2, because `s$2_3$' performs better than the others in (a). (d)-(f) are also in the same way. It is obvious that separating BN layers significantly outperforms separating the entire block, which motivates us to explore more BN separation schemes.
}
\label{fig1}
\end{figure*}

\paragraph{Neural Architecture Search.}
Existing NAS works can be grouped into micro search methods and macro search methods \cite{hu2020tf}.
The micro search methods aim to design robust cells and then stack these cells to constitute the neural architecture.
Traditional methods mainly depend on evolutionary algorithms or reinforcement learning to tackle the discrete search problem \cite{real2019regularized,zoph2018learning}.
Recently, DARTS \cite{liu2018darts} first proposes a differentiable search strategy, which greatly reduces the computational overhead compared with the traditional methods \cite{dong2019searching,xu2019pc,chen2019progressive}.
The macro search methods search the whole neural architecture, which is thought to be more flexible than searching cells \cite{cai2018proxylessnas,tan2019efficientnet}.
\cite{baker2016designing} introduces Q-learning to select layers sequentially.
\cite{wu2019fbnet} searches accurate and efficient architectures for mobile devices.
However, these methods are all designed for single-modality tasks, with no need for considering the modality discrepancy.

\section{Method}

\begin{figure}[t]
\centering
\includegraphics[width=0.42 \textwidth]{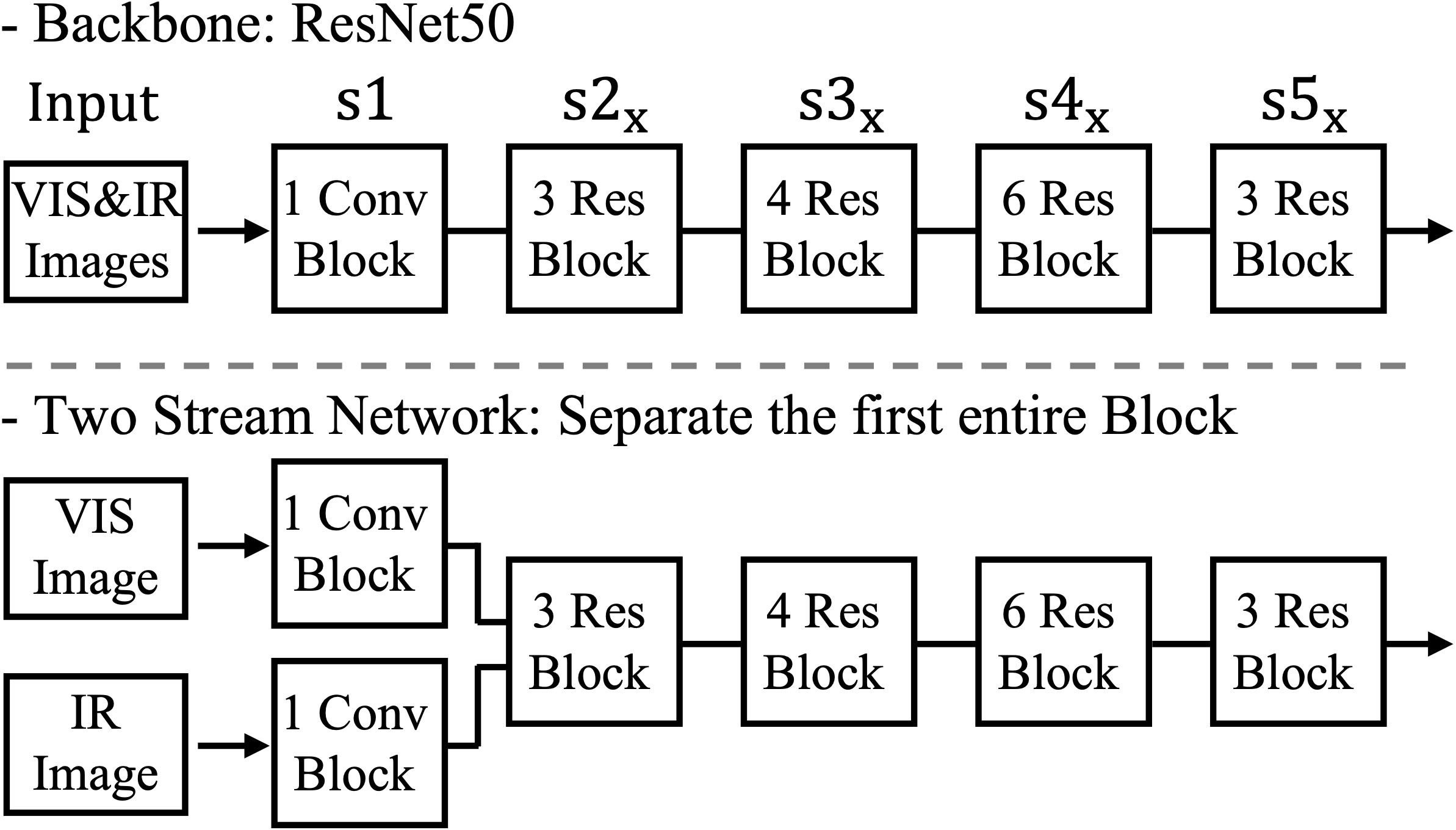}
\caption{\textbf{Top:} The architecture of the backbone ResNet50 \cite{he2016deep} that includes 5 stages. `s$2_\text{x}$' denotes the x-th ResBlock in stage2. A ConvBlock has one convolution layer, one BN layer \cite{ioffe2015batch} and one ReLU layer. A ResBlock contains a total of three convolution layers, three BN layers and three ReLU layers. Please refer to \cite{he2016deep} for details.
\textbf{Bottom:} An example of the two stream network, where the first entire block is separated into two branches for VIS and IR images, respectively.
}
\label{fig-resnet50}
\end{figure}

\subsection{Analyses of Manually Designed Architectures} \label{rethinking}
For fair comparisons with other VI-ReID methods, we employ ResNet50\footnote{Other networks are also applicable.} pre-trained on ImageNet as the backbone, whose architecture is reported in Fig.~\ref{fig-resnet50}. There are 5 stages in ResNet50, where each stage consists of blocks, including ConvBlock and ResBlock.
As mentioned in Section~\ref{Introduction}, due to the absence of consensus on the optimal architecture, we perform systematic research to assess the impact of different architecture designs.
In general, the architecture is designed in units of the entire block \cite{ye2019bi,ye2020dynamic}, such as separating one or five blocks into two branches \cite{ye2018visible,liu2020enhancing}.
Besides, we also design the architecture in units of the BN layers in the block, \textit{i.e.} only separating the BN layers rather than all the layers in the block.
This is inspired by \cite{xie2020adversarial}, which reveals that separating BN layers for the data from different domains outperforms sharing the BN layers.

First of all, we evaluate the performance of separating one block, including separating all the layers in the block and solely separating the BN layers in the block.
Concretely, we separate the blocks in ResNet50 one by one to learn modality-specific representations, and share the remaining to learn modality-sharable representations.
Fig.~\ref{fig1} (a) depicts the results of all the potential architectures as well as the result of one baseline: sharing the whole architecture without separation.
We have three observations from Fig.~\ref{fig1} (a):
(1) the baseline generally performs worse than separating one block, suggesting the necessity of separating blocks to learn modality-specific representations;
(2) separating different blocks yields much different performances. For example, when separating the entire block, `s$4_1$' and `s$5_1$' lead to the best and the worst results, respectively. This implies that we need to carefully treat each layer in the design process;
(3) separating the BN layers in the block is more suitable than separating the entire block, since the former (the red line) gains much better results than the latter (the blue line) in all separation schemes.
Subsequently, we further separate two blocks for each time.
Fig.~\ref{fig1} (b)-(f) display the results when we first fixedly separate a block in stage1, stage2, stage3, stage4 and stage5 respectively, and then traverse through the remaining stages to separate the other one.
The performance of separating one block in the first step is also reported as a baseline result.
A distinct observation from these results is that separating two blocks generally outperforms than separating a single one, especially when only separating the BN layers in the block.

With these observations, we arrive at a conclusion that appropriately separating BN layers can lead to better performances.
Consequently, the essential objective is to find the optimal separation scheme for each BN layer in the backbone.
As mentioned in Section~\ref{Introduction}, since it is intractable to manually traverse through all potential architectures, we develop a novel CM-NAS to automatically find the best one.

\subsection{Cross-Modality NAS} \label{search space}
\begin{figure}[t]
\centering
\includegraphics[width=0.47 \textwidth]{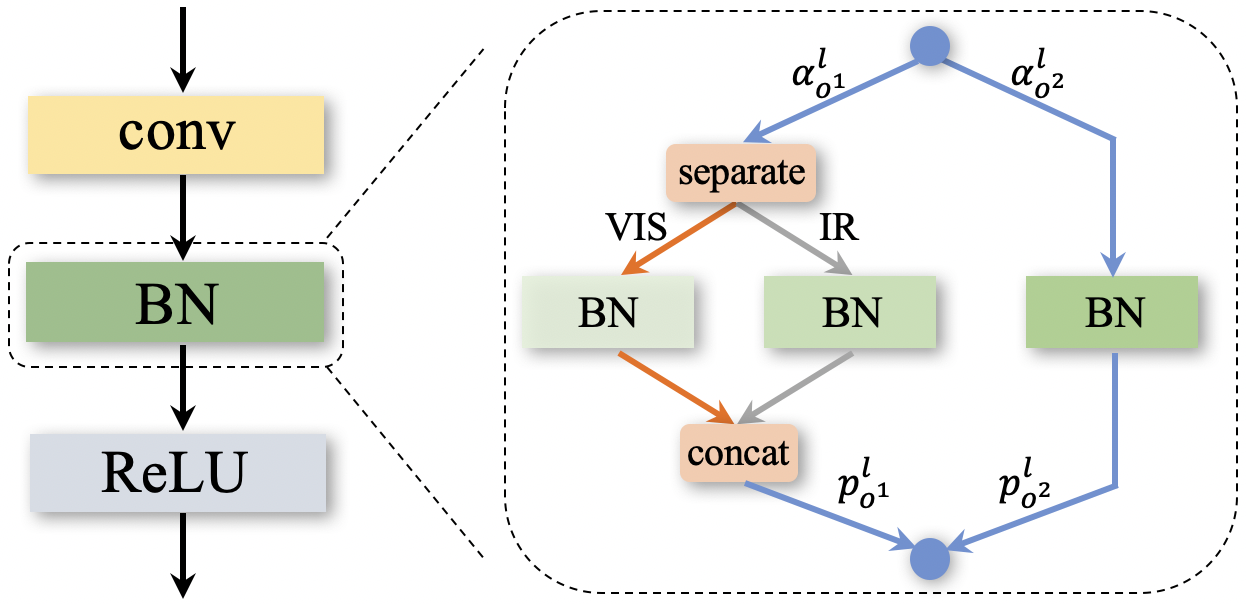}
\caption{BN-oriented search space of CM-NAS. All BN layers in the backbone are reshaped as searchable units. Each BN layer has two candidate operations: employing two separate parameters for VIS and IR data respectively, or sharing its parameters for both the two modalities of data.
}
\label{fig2}
\end{figure}

\paragraph{Search Space.}
In view of the above analyses, our architecture design revolves around which BN layers should be separated and which BN layers should be shared.
With this in mind, we design a BN-oriented search space that is depicted in Fig.~\ref{fig2}. In our search space, all BN layers in the backbone are reshaped as searchable units, and each BN layer has two candidate operations: employing separate or shared parameters. If a BN layer chooses the former, this BN layer will have two separate parameters that are learned from VIS and IR data, respectively. Otherwise, this BN layer will share parameters that are learned from both the two modalities of data.

Formally, let $o^1$ and $o^2$ denote the above two candidate operations, respectively.
In each BN layer $l$, we assign an architecture parameter $\alpha_{o^1}^l$ to the operation $o^1$ and the other architecture parameter $\alpha_{o^2}^l$ to the remaining operation $o^2$.
When $\alpha_{o^1}^l$ = 1 and $\alpha_{o^2}^l$ = 0, it means that the BN layer $l$ uses separate parameters.
Otherwise, when $\alpha_{o^1}^l$ = 0 and $\alpha_{o^2}^l$ = 1, the BN layer $l$ shares its parameters.
In practice, instead of searching on such discrete candidate architectures, we relax the search space to make it can be optimized via gradient descent \cite{liu2018darts}.
Concretely, we relax the binary architecture parameters $\alpha_{o^i}^l$ ($i\in\{1, 2\}$) to be continuous, and then obtain the probability of choosing the corresponding operation by computing a softmax over all architecture parameters:
\begin{equation}\label{eq1}
    p_{o^i}^l = \frac{\text{exp}(\alpha_{o^i}^l)}{\text{exp}(\alpha_{o^1}^l) + \text{exp}(\alpha_{o^2}^l)}.
\end{equation}
The larger the value of $p_{o^i}^l$, the more likely the BN layer $l$ is to choose the operation $o^i$.
The output of the BN layer $l$ is a weighted sum of all possible operations:
\begin{equation}\label{eq2}
    x_{l+1} = p_{o^1}^l \cdot o_1(x_l) + p_{o^2}^l \cdot o_2(x_l),
\end{equation}
where $o_i(x_l)$ denotes that the operation $o_i$ is applied to the input $x_l$.
In such a case, the search process is transformed into the learning of a set of architecture parameters $\alpha = \{ \alpha_{o^i}^l \}$.
Furthermore, since the network weights $w$, such as the weights of convolution layers, also need to be learned, we are required to tackle the following bi-level optimization problem \cite{liu2018darts,dong2019searching,wu2019fbnet}:
\begin{equation}\label{eq3}
\begin{split}
    & \min_{\alpha} \mathcal{L}_{\text{val}}(w^*, \alpha), \\
    & s.t. \quad w^* = \arg \min_{w}\mathcal{L}_{\text{train}}(w, \alpha).
\end{split}
\end{equation}
The goal of Eq.~(\ref{eq3}) is to search architecture parameters $\alpha^*$ that minimize the validation loss $\mathcal{L}_{\text{val}}(w^*, \alpha^*)$, where the network weights $w^*$ are obtained by minimizing the training loss $\mathcal{L}_{\text{train}}(w, \alpha)$.
After training, for each BN layer $l$, we choose the operation with a larger probability and abandon the other one, yielding a discrete architecture.
For instance, when $p_{o^1}^l > p_{o^2}^l$, we will choose the operation $o_1$, \textit{i.e.} employing two separate parameters in the BN layer $l$.
In addition, it is obvious that the training and the validation losses play critical roles in the search process, which are introduced in the following contents.

\paragraph{Objective Function.}
Up to now, the remaining part of our CM-NAS is to design an appropriate objective function to better guide the cross-modality search process.
To begin with, Class-specific Maximum Mean Discrepancy (CMMD) \cite{gretton2007kernel,kang2019contrastive} is a commonly used measure for the modality discrepancy:
\begin{equation}\label{eq6}
    \mathcal{L}_\text{CMMD} = \frac{1}{C} \sum_{c=1}^C \left|\left|\frac{1}{m^c} \sum_{i=1}^{m^c} \psi(f_i^{c,vis}) -  \frac{1}{n^c} \sum_{j=1}^{n^c} \psi(f_j^{c,ir}) \right|\right|_{\mathcal{H}},
\end{equation}
where $f^{c, vis}$ and $f^{c, ir}$ denote the features of VIS and IR images belonging to the $c$-th class, respectively.
$m^c$ and $n^c$ are the numbers of the corresponding features.
$\psi(\cdot)$ is a function that maps features into a universal Reproducing Kernel Hilbert Space (RKHS) \cite{steinwart2001influence}.
We construct $\psi(\cdot)$ via a polynomial kernel $k$($x$, $y$) = $\psi$($x$)$ \cdot \psi$($y$) = ($x \cdot y$)$^2$.

Subsequently, given that the importance of feature correlations \cite{peng2019correlation,tung2019similarity}, we also constrain the correlation consistency between the features of the VIS and IR modalities.
Formally, let $\mathbf{F}_{vis}$ and $\mathbf{F}_{ir}$ be the set of embedding features of the VIS and the IR data respectively:
\begin{equation}\label{eq7}
\begin{split}
    & \mathbf{F}_{vis} = \text{matrix}(f_1^{vis}, f_2^{vis},..., f_n^{vis}), \\
    & \mathbf{F}_{ir} = \text{matrix}(f_1^{ir}, f_2^{ir},..., f_n^{ir}).
\end{split}
\end{equation}
In practice, we sample $n$ VIS and $n$ IR images for each time, because constraining the correlation consistency requires the same number of data \cite{tung2019similarity}. Meanwhile, $f_i^{vis}$ and $f_i^{ir}$ ($i \in \{1,...,n\}$) belong to the same identity.
Correlation matrices $\tilde{\mathbf{G}}_{vis}$ and $\tilde{\mathbf{G}}_{ir}$ that reflect pairwise similarities among the features are given by:
\begin{equation}\label{eq8}
\begin{split}
    \tilde{\mathbf{G}}_{vis} = \mathbf{F}_{vis} \cdot \mathbf{F}_{vis}^{\top}, \quad \tilde{\mathbf{G}}_{ir} = \mathbf{F}_{ir} \cdot \mathbf{F}_{ir}^{\top}.
\end{split}
\end{equation}
Then, a row-wise $L_2$ normalization is applied on the two correlation matrices:
\begin{equation}\label{eq9}
\begin{split}
    \mathbf{G}_{vis[i,:]} = \frac{\tilde{\mathbf{G}}_{vis[i,:]}}{|| \tilde{\mathbf{G}}_{vis[i,:]} ||_2}, \quad \mathbf{G}_{ir[i,:]} = \frac{\tilde{\mathbf{G}}_{ir[i,:]}}{|| \tilde{\mathbf{G}}_{ir[i,:]} ||_2},
\end{split}
\end{equation}
where the notation $[i,:]$ denotes the $i$-th row in a matrix.
A correlation consistency loss $\mathcal{L}_\text{CC}$ is further developed to penalize the difference between $\mathbf{G}_{vis}$ and $\mathbf{G}_{ir}$:
\begin{equation}\label{eq10}
    \mathcal{L}_\text{CC} = \frac{1}{n^2} || \mathbf{G}_{vis} - \mathbf{G}_{ir} ||^2_F,
\end{equation}
where $||\cdot||_F$ denotes a Frobenius norm.

The weighted sum of the above $\mathcal{L}_\text{CMMD}$ and $\mathcal{L}_\text{CC}$ is named as $\mathcal{L}_\text{C3MMD}$:
\begin{equation}\label{eq11}
    \mathcal{L}_\text{C3MMD} = \lambda_1 \mathcal{L}_\text{CMMD} + \lambda_2 \mathcal{L}_\text{CC},
\end{equation}
where $\lambda_1$ and $\lambda_2$ are trade-off parameters.
Finally, in addition to $\mathcal{L}_\text{C3MMD}$, our objective function also includes a basic loss $\mathcal{L}_\text{basic}$ \cite{wang2020cross,ye2020dynamic} that consists of a classification loss $\mathcal{L}_\text{cls}$ and a triplet loss $\mathcal{L}_\text{triplet}$ to learn embedding features.
As a result, the training loss $\mathcal{L}_\text{train}$ in Eq.~(\ref{eq3}) is the combination of $\mathcal{L}_\text{basic}$ and $\mathcal{L}_\text{C3MMD}$:
\begin{equation}\label{eq12}
    \mathcal{L}_\text{train} = \mathcal{L}_\text{basic} + \mathcal{L}_\text{C3MMD}.
\end{equation}
The validation loss $\mathcal{L}_\text{val}$ in Eq.~(\ref{eq3}) has the same form as the training loss.

\section{Experiment}
\subsection{Experimental Settings}
\paragraph{Dataset.}
SYSU-MM01 \cite{wu2017rgb} is a pioneer benchmark for the research of VI-ReID.
This dataset consists of a total of 287,628 VIS images taken by 4 VIS cameras in the daytime, and 15,792 near-IR images taken by 2 near-IR cameras in the dark environment.
These images are captured in both indoor and outdoor scenarios with abundant poses and viewpoints.
The training set and the testing set have 395 and 96 person identities, respectively.
Following \cite{wu2017rgb}, there are two testing models: \emph{all-search} and \emph{indoor-search}.
For the former, the gallery set contains VIS images in both indoor and outdoor scenarios.
For the latter, the gallery set merely contains VIS images in the indoor scenario.
Besides, for both models, there are also two settings: \emph{single-shot} and \emph{multi-shot}.
It means 1 or 10 VIS images of a person identity are randomly chosen to constitute the gallery set.

RegDB \cite{nguyen2017person} is built by a dual camera acquisition system that includes a VIS camera and a thermal-IR camera.
The two cameras are attached together to take photos at the same time, acquiring a total of 4,120 paired VIS-IR images from 412 person identities (each identity has 10 VIS images and 10 thermal-IR images).
Following \cite{ye2018hierarchical}, 2,060 images from 206 person identities are randomly chosen as the training set and the remaining 2,060 images from 206 identities constitute the testing set.
There are two evaluation settings: \emph{Visible to Infrared} and \emph{Infrared to Visible}. Take the former for example, it denotes leveraging the VIS images as the probe set and the IR images as the gallery set.

\paragraph{Evaluation Metrics.}
Following existing VI-ReID methods \cite{wang2019rgb,ye2019bi}, we adopt Cumulative Matching Characteristic (CMC) and mean Average Precision (mAP) as evaluation metrics.
Moreover, the reported result on the SYSU-MM01 dataset is an average performance of 10 times repeated random probe/gallery splits \cite{wu2017rgb}, while that on the RegDB dataset is an average performance of 10 times random training/testing splits \cite{ye2018hierarchical,wang2019learning}.

\begin{table*}[t]
    \centering
    \caption{Evaluations of CM-NAS on the SYSU-MM01 dataset under the \emph{all-search} setting and the RegDB dataset. R1, R10 and R20 denote Rank-1, Rank-10 and Rank-20 accuracies (\%), respectively. mAP denotes the mean average precision score (\%). phase1 and phase2 correspond to the searching and the training phases of NAS, respectively. IN stands for Instance Normalization \cite{ulyanov2016instance,jin2020style}.
    }
    \label{table-1}
    \resizebox{0.985\textwidth}{!} {
    \begin{tabular}{l|c|cccc|cccc"cccc|cccc}
        \myline
        \multirow{3}{*}{Method} & \multirow{3}{*}{$\mathcal{L}_\text{C3MMD}$} & \multicolumn{8}{c"}{SYSU-MM01 (All-Search)} & \multicolumn{8}{c}{RegDB} \\
        \cline{3-18}
        &  & \multicolumn{4}{c|}{Single-Shot} & \multicolumn{4}{c"}{Multi-Shot} & \multicolumn{4}{c|}{Visible to Infrared} & \multicolumn{4}{c}{Infrared to Visible} \\
        \cline{3-18}
        & & R1 & R10 & R20 & mAP & R1 & R10 & R20 & mAP & R1 & R10 & R20 & mAP & R1 & R10 & R20 & mAP  \\
        \hline
        one-stream            & \ding{55} & 54.49 & 89.85 & 95.78 & 53.32 & 61.26 & 85.22 & 92.20 & 46.53 & 75.81 & 90.57 & 95.07 & 71.97 & 74.03 & 90.30 & 94.43 & 70.01 \\
        one-stream (IN)       & \ding{55} & 45.54 & 86.51 & 94.20 & 44.48 & 52.37 & 81.51 & 96.46 & 36.76 & 67.48 & 85.47 & 91.15 & 63.75 & 63.62 & 83.01 & 89.23 & 60.08 \\
        two-stream            & \ding{55} & 56.20 & 90.62 & 95.83 & 54.22 & 62.70 & 93.44 & 97.61 & 46.87 & 76.93 & 91.24 & 95.39 & 73.02 & 74.51 & 90.46 & 94.66 & 70.55 \\
        two-stream (IN)       & \ding{55} & 46.01 & 86.83 & 94.36 & 45.06 & 52.71 & 90.32 & 96.21 & 37.60 & 68.39 & 85.63 & 91.30 & 64.50 & 64.04 & 83.56 & 89.32 & 61.28 \\
        two-stream (BN)       & \ding{55} & 56.94 & 90.77 & 96.18 & 55.24 & 62.74 & 92.81 & 96.93 & 47.91 & 77.88 & 91.96 & 95.74 & 73.98 & 75.00 & 90.71 & 94.56 & 71.64 \\
        Auto-ReID \cite{quan2019auto} & \ding{55} & 43.08 & 85.88 & 93.43 & 42.83 & 51.67 & 88.55 & 95.87 & 35.01 & 63.07 & 83.44 & 91.34 & 61.08 & 62.72 & 82.32 & 90.68 & 59.09 \\
        search                  & \ding{55} & 59.56 & 91.43 & 96.24 & 56.79 & 66.12 & 94.21 & 97.90 & 49.89 & 78.77 & 92.94 & 96.29 & 76.05 & 77.55 & 92.44 & 95.87 & 74.86 \\
        \hline
        one-stream           & \ding{51} & 56.18 & 90.80 & 96.05 & 54.47 & 63.36 & 93.15 & 97.53 & 47.60 & 77.42 & 91.46 & 95.32 & 74.57 & 75.45 & 90.67 & 94.66 & 71.84 \\
        one-stream (IN)       & \ding{51} & 46.93 & 87.10 & 94.59 & 45.83 & 54.43 & 90.23 & 97.11 & 38.51 & 70.96 & 88.34 & 93.08 & 66.35 & 68.24 & 86.18 & 91.92 & 62.76 \\
        two-stream              & \ding{51} & 57.86 & 90.72 & 96.21 & 55.97 & 64.28 & 93.66 & 97.97 & 48.41 & 78.03 & 91.79 & 95.37 & 74.96 & 75.52 & 90.77 & 94.68 & 72.04 \\
        two-stream (IN)       & \ding{51} & 47.38 & 88.66 & 95.74 & 46.78 & 55.18 & 92.65 & 97.75 & 39.57 & 71.35 & 88.62 & 93.35 & 66.94 & 68.94 & 86.30 & 92.05 & 63.72 \\
        two-stream (BN)         & \ding{51} & 57.94 & 91.70 & 96.76 & 56.30 & 65.05 & 93.58 & 97.88 & 49.33 & 78.48 & 92.65 & 96.27 & 75.53 & 75.87 & 91.10 & 94.64 & 73.10 \\
        Auto-ReID \cite{quan2019auto} & \ding{51} & 44.71 & 86.10 & 93.91 & 44.58 & 53.41 & 88.85 & 96.31 & 36.63 & 65.52 & 83.80 & 91.47 & 62.70 & 64.64 & 83.13 & 90.94 & 60.56 \\
        \hline
        search                  & phase1    & 59.55 & 91.83 & 96.93 & 57.62 & 66.72 & 94.52 & 97.63 & 50.73 & 79.71 & 93.25 & 96.38 & 76.65 & 78.20 & 92.70 & 96.04 & 75.28 \\
        search                  & phase2  & 59.95 & 91.74 & 96.80 & 57.72 & 66.93 & 94.55 & 97.96 & 51.02 & 82.36 & 94.16 & 97.27 & 78.83 & 81.20 & 93.63 & 96.54 & 77.14 \\
        search                  & phases1\&2& \bf{61.99} & \bf{92.87} & \bf{97.25} & \bf{60.02} & \bf{68.68} & \bf{94.92} & \bf{98.36} & \bf{53.45} & \bf{84.54} & \bf{95.18} & \bf{97.85} & \bf{80.32} & \bf{82.57} & \bf{94.51} & \bf{97.37} & \bf{78.31} \\
        \hline
        search  & w/o $\mathcal{L}_\text{CMMD}$ & 60.77 & 91.73 & 96.46 & 58.74 & 67.56 & 94.69 & 98.14 & 51.93 & 82.18 & 94.41 & 97.69 & 78.92 & 81.53 & 93.98 & 96.86 & 77.45 \\
        search  & w/o $\mathcal{L}_\text{CC}$   & 60.27 & 91.96 & 96.77 & 58.45 & 67.25 & 94.50 & 98.09 & 51.39 & 83.75 & 94.86 & 97.84 & 79.72 & 81.87 & 94.29 & 97.09 & 77.45 \\
        search  & $\mathcal{L}_\text{CMMD} \to \mathcal{L}_\text{MMD}$ & 60.83 & 92.14 & 96.79 & 58.92 & 67.99 & 94.76 & 97.90 & 52.37 & 82.79 & 95.06 & 97.74 & 79.25 & 81.68 & 94.06 & 96.91 & 77.58 \\
        \myline
    \end{tabular}}
\end{table*}

\paragraph{Implementation Details.}
We adopt ResNet50 \cite{he2016deep} pre-trained on ImageNet as the backbone. 
The input images are first padded with 10 and then randomly cropped to 256$\times$128.
Random horizontal flipping and random erasing are further imposed as data augmentation.
Adam ($\beta_1$=0.5, $\beta_2$=0.999) is employed as an optimizer with 5e-4 weight decay.
The initial learning rate is set to 0.01 and divided by 10 at the 40-th and the 70-th epochs. The training process is finished at the 120-th epoch.
One training batch contains 8 identities and each identity has 4 VIS images as well as 4 IR images.
These experimental settings mainly refer to previous works \cite{lu2020cross,luo2019bag}.
The trade-off parameters $\lambda_1$ and $\lambda_2$ in Eq.~(\ref{eq11}) are set to 5.0 and 0.05, respectively.
Following the standard process of NAS \cite{liu2018darts,fang2020densely,hu2020tf}, there are two phases to train the network.
In the first phase, all BN layers in the backbone are set to be searchable and Eq.~(\ref{eq3}) is utilized as the loss function, aiming to find the optimal neural architecture.
Since the search process needs a validation set, we divide the original training set into a new training set and a validation set by a ratio of identities of 8:2.
In the second phase, the architecture of the network is fixed to the best one and we retrain the network with Eq.~(\ref{eq12}) for the evaluation of VI-ReID.
The training set in the second phase is the original one without splitting the validation set.

\subsection{Experimental Analyses} \label{experimental analyses}
\paragraph{Evaluation of the Search Space.}
We evaluate the search space proposed in Section~\ref{search space} via a series of comparisons.
Concretely, we only employ the basic loss $\mathcal{L}_\text{basic}$ in Eq.(\ref{eq12}) as the objective function.
The compared six baselines are:
(1) \emph{one-stream} that shares the whole architecture for both VIS and IR modalities;
(2) \emph{one-stream (IN)} that replaces all BN in \emph{one-stream} with IN, since the latter has the potential to reduce modality discrepancy \cite{jin2020style};
(3) \emph{two-stream} that separates blocks of stage1 and stage2 to learn modality-specific representations, and shares remaining blocks to learn modality-sharable representations;
(4) \emph{two-stream (IN)} that replaces all BN in \emph{two-stream} with IN;
(5) \emph{two-stream (BN)} that only separates BN layers rather than all layers in the stage1 and stage2 blocks;
(6) \emph{Auto-ReID} \cite{quan2019auto} that searches the whole architecture rather than merely BN layers.
To make a fair comparison, we constrain the searched architecture of Auto-ReID to have the same FLOPs as ResNet50.

The comparison results on the SYSU-MM01 and the RegDB datasets are reported Table~\ref{table-1}, from which we have five observations.
First, replacing BN with IN leads to performance degradation. This is because that although IN can minimize modality discrepancy by shifting style, it also brings the loss of discriminative information \cite{jin2020style}.
Second, the two-stream method performs better than the one-stream method, implying the necessity of separating blocks.
Third, only separating the BN layers in the block is superior than separating the entire block, which indicates the significance of the BN layer in learning cross-modality representations.
The above two observations are consistent with those in Section~\ref{rethinking}, which motivate us to develop a BN-oriented search algorithm to automatically decide the separation of BN layers.
Fourth, the performances of Auto-ReID are only comparable with the two stream method.
This is because that the search space of Auto-ReID, which is specially designed for the single-modality task, is powerless to bridge the modality discrepancy in VI-ReID.
Fifth, our search algorithm outperforms all the competitors.
For instance, the search algorithm exceeds the BN-oriented two-stream method by 2.62\% and 1.55\% in terms of the Rank-1 and mAP on the SYSU-MM01 dataset under the \emph{single-shot}$\&$\emph{all-search} setting.
This demonstrates that our search algorithm can indeed find a more suitable neural architecture.
Besides, compared with Auto-ReID \cite{quan2019auto}, our method improves the Rank-1/mAP by 16.48\%/13.96\% on SYSU-MM01 and 15.70\%/14.97\% on RegDB.
The unsatisfactory performances of Auto-ReID may be because that its cell-based search space is powerless to bridge the modality discrepancy.
This further verifies the merit of our specially designed BN-oriented search space.

\paragraph{Evaluation of the Objective Function.}
In order to verify the effectiveness of $\mathcal{L}_\text{C3MMD}$ in Eq.~(\ref{eq11}), we elaborately design several comparative experiments.
Specifically, we add $\mathcal{L}_\text{C3MMD}$ to the training of the aforementioned seven methods, including the six baselines as well as the search method.
Since the search method has two training phases, there are three manners to add $\mathcal{L}_\text{C3MMD}$:
(1) only adding $\mathcal{L}_\text{C3MMD}$ in the first phase, which means that the first phase is trained with both $\mathcal{L}_\text{basic}$ and $\mathcal{L}_\text{C3MMD}$ while the second phase is solely trained with $\mathcal{L}_\text{basic}$;
(2) only adding $\mathcal{L}_\text{C3MMD}$ in the second phase;
(3) adding $\mathcal{L}_\text{C3MMD}$ in the both two phases.
The results of the above methods are listed Table~\ref{table-1}.
Regarding to the six baselines, compared with only employing $\mathcal{L}_\text{basic}$ (the first six rows in Table~\ref{table-1}), it is observed that their performances are all improved after adding $\mathcal{L}_\text{C3MMD}$.
For example, on the SYSU-MM01 dataset under the \emph{single-shot}$\&$\emph{all-search} setting, the Rank-1 accuracy and the mAP score of the BN-oriented two-stream method increase by 1.0\% and 1.06\%, respectively.
Such improvements adequately reveal the effectiveness of the proposed C3MMD loss.
For the search method, we can see that adding $\mathcal{L}_\text{C3MMD}$ in the both two phases performs better than adding it in only one single phase.

Furthermore, we also investigate the effect of the two components in $\mathcal{L}_\text{C3MMD}$, \textit{i.e.} $\mathcal{L}_\text{CMMD}$ and $\mathcal{L}_\text{CC}$.
The results of the ablation study are shown in the bottom three rows of Table~\ref{table-1}.
We find that removing any component will lead to performance degradation, suggesting the importance of both $\mathcal{L}_\text{CMMD}$ and $\mathcal{L}_\text{CC}$.
Furthermore, replacing $\mathcal{L}_\text{CMMD}$ with $\mathcal{L}_\text{MMD}$ \cite{gretton2007kernel} also results in inferior performances.
This is because that compared with the latter, the former takes class labels into consideration and thus focuses on more specific modality distributions.

\paragraph{Parameter Analyses.}
We analyze the two trade-off parameters $\lambda_1$ and $\lambda_2$ in Eq.~(\ref{eq11}) on SYSU-MM01. 
The Rank-1 and the mAP results of CM-NAS with different $\lambda_1$ and $\lambda_2$ are exhibited in Fig.~\ref{fig4}.
We can see that our method is not sensitive to the parameters $\lambda_1$ and $\lambda_2$ in a large range.
For instance, when $\lambda_1$ changes from 0.01 to 0.1, the Rank-1 accuracy only changes 0.67\%.
Moreover, the performance drops when the parameters are set too large, such as $\lambda_1$ = 0.2 or $\lambda_2$ = 15.
The most suitable parameter setting is that $\lambda_1$ = 0.05 and $\lambda_2$ = 5.0.

\begin{figure}[b]
\centering
\includegraphics[width=0.475 \textwidth]{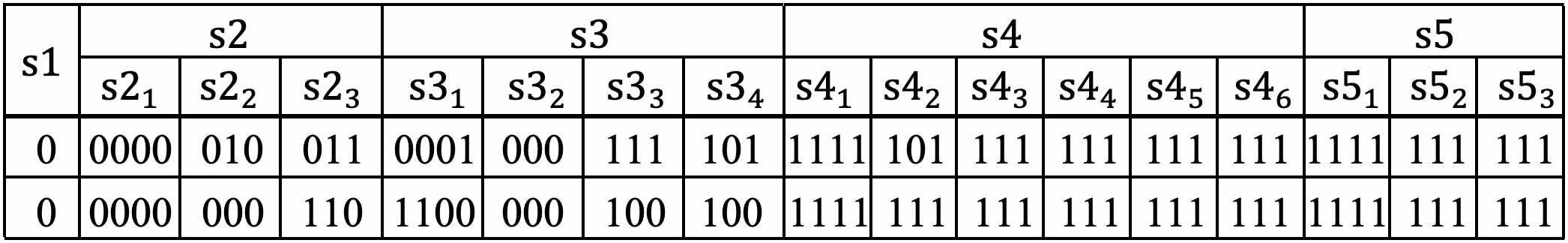}
\caption{Searched architectures on SYSU-MM01 (top) and RegDB (bottom). Each bit denotes the state of the corresponding BN layer, where `0' means separating and `1' means sharing. The symbol like `s$2$' denotes the architecture name of ResNet50, as shown in Fig.~\ref{fig-resnet50}.
}
\label{fig-arc}
\end{figure}

\paragraph{Architecture Analyses.}
The searched architectures on SYSU-MM01 and RegDB are depicted in Fig.~\ref{fig-arc}.
First, it is observed that compared with low-level BN layers, high-level BN layers prefer to share parameters, which is in line with manual experience.
Second, there are also some low-level BN layers choose sharing and some high-level BN layers choose separating.
Compared with previous methods that directly separate the layers in the first one \cite{ye2020cross,ye2020dynamic} or five \cite{ye2018visible} blocks, our searched architecture is such complex and is hard to be manually designed, which suggests the necessity of the search algorithm.
Third, all BN layers in stage5 are shared, which is consistent with the observation in Fig.~\ref{fig1} (f).
That is, separating the layers of stage5 always results in inferior performances.

\begin{figure}[t]
\centering
\includegraphics[width=0.45 \textwidth]{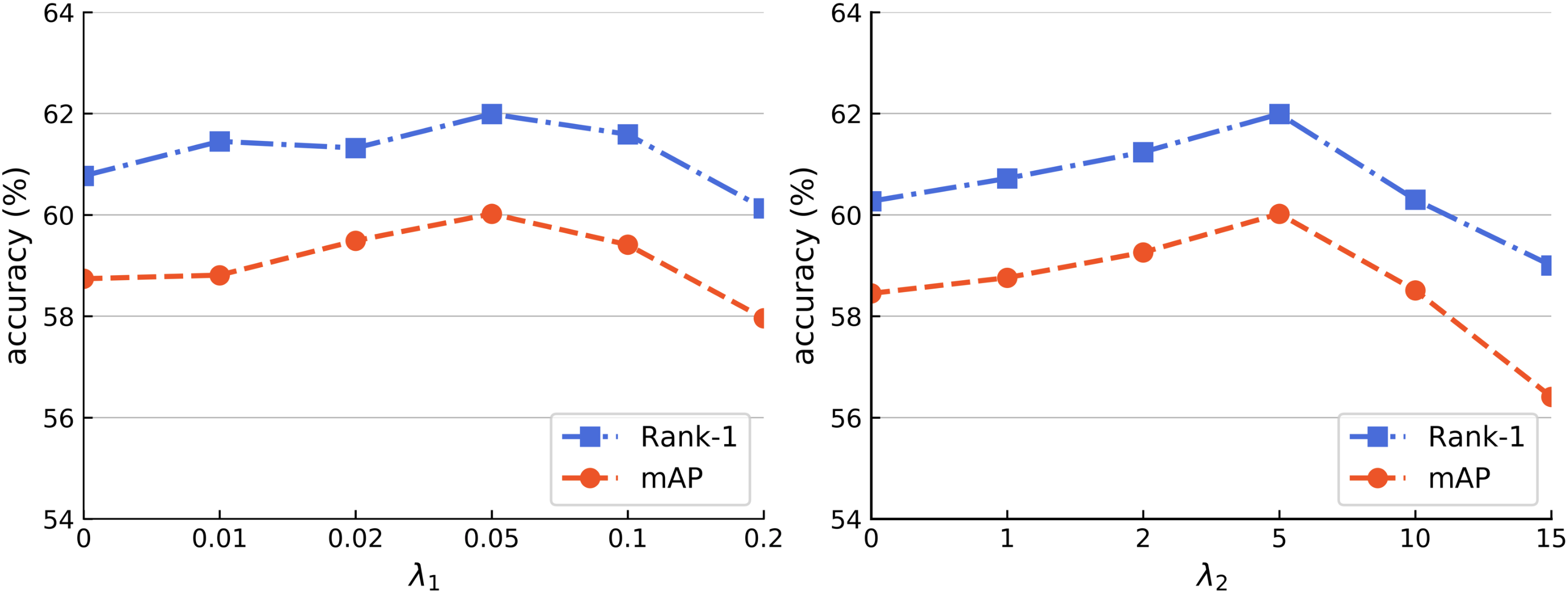}
\caption{Parameter analyses of $\lambda_1$ and $\lambda_2$ in Eq.~(\ref{eq11}) on the SYSU-MM01 dataset under the \emph{single-shot}$\&$\emph{all-search} setting. We first fix $\lambda_2$ to 5.0 and vary the value of $\lambda_1$ (left). Then, we fix $\lambda_1$ to 0.05 and vary the value of $\lambda_2$ (right). Rank-1 accuracies and mAP scores are reported for comparisons.
}
\label{fig4}
\end{figure}

\begin{table*}[t]
    \centering
    \caption{Comparisons with state-of-the-art methods on the SYSU-MM01 dataset.
    cm-SSFT$^*$ denotes that we report the results of cm-SSFT under the single-query setting \cite{lu2020cross} for fair comparisons with other methods.
    CM-NAS$^{\dagger}$ means first searching a neural architecture on the RegDB dataset and then training the searched neural network on the SYSU-MM01 dataset with the loss function Eq.~(\ref{eq12}).}
    \label{table-2}
    \resizebox{0.985\textwidth}{!} {
    \begin{tabular}{l|c|cccc|cccc|cccc|cccc}
        \myline
        \multirow{3}{*}{Method} & \multirow{3}{*}{Venue} & \multicolumn{8}{c|}{All-Search} & \multicolumn{8}{c}{Indoor-Search} \\
        \cline{3-18}
        &  & \multicolumn{4}{c|}{Single-Shot} & \multicolumn{4}{c|}{Multi-Shot} & \multicolumn{4}{c|}{Single-Shot} & \multicolumn{4}{c}{Multi-Shot} \\
        \cline{3-18}
        & & R1 & R10 & R20 & mAP & R1 & R10 & R20 & mAP & R1 & R10 & R20 & mAP & R1 & R10 & R20 & mAP  \\
        \hline
        Zero-Pad~\cite{wu2017rgb} & ICCV-17 & 14.80 & 54.12 & 71.33 & 15.95 & 19.13 & 61.40 & 78.41 & 10.89 & 20.58 & 68.38 & 85.79 & 26.92 & 24.43 & 75.86 & 91.32 & 18.64 \\
        TONE~\cite{ye2018hierarchical} & AAAI-18 & 12.52 & 50.72 & 68.60 & 14.42 & - & - & - & - & 20.82 & 68.86 & 84.46 & 26.38 & - & - & - & - \\
        HCML~\cite{ye2018hierarchical} & AAAI-18 & 14.32 & 53.16 & 69.17 & 16.16 & - & - & - & - & 24.52 & 73.25 & 86.73 & 30.08 & - & - & - & - \\
        cmGAN~\cite{dai2018cross} & IJCAI-18 & 26.97 & 67.51 & 80.56 & 27.80 & 31.49 & 72.74 & 85.01 & 22.27 & 31.63 & 77.23 & 89.18 & 42.19 & 37.00 & 80.94 & 92.11 & 32.76 \\
        BDTR~\cite{ye2019bi} & IJCAI-18 & 27.32 & 66.96 & 81.07 & 27.32 & - & - & - & - & 31.92 & 77.18 & 89.28 & 41.86 & - & - & - & - \\
        eBDTR~\cite{ye2019bi} & TIFS-19 & 27.82 & 67.34 & 81.34 & 28.42 & - & - & - & - & 32.46 & 77.42 & 89.62 & 42.46 & - & - & - & - \\
        HSME~\cite{hao2019hsme} & AAAI-19 & 20.68 & 62.74 & 77.95 & 23.12 & - & - & - & - & - & - & - & - & - & - & - & - \\
        D$^2$RL~\cite{wang2019learning} & CVPR-19 & 28.9 & 70.6 & 82.4 & 29.2 & - & - & - & - & - & - & - & - & - & - & - & - \\
        MSR~\cite{feng2019learning} & TIP-19 & 37.35 & 83.40 & 93.34 & 38.11 & 43.86 & 86.94 & 95.68 & 30.48 & 39.64 & 89.29 & 97.66 & 50.88 & 46.56 & 93.57 & 98.80 & 40.08 \\
        AlignGAN~\cite{wang2019rgb} & ICCV-19 & 42.4 & 85.0 & 93.7 & 40.7 & 51.5 & 89.4 & 95.7 & 33.9 & 45.9 & 87.6 & 94.4 & 54.3 & 57.1 & 92.7 & 97.4 & 45.3 \\
        JSIA-ReID~\cite{wang2020cross} & AAAI-20 & 38.1 & 80.7 & 89.9 & 36.9 & 45.1 & 85.7 & 93.8 & 29.5 & 43.8 & 86.2 & 94.2 & 52.9 & 52.7 & 91.1 & 96.4 & 42.7 \\
        Xmodal~\cite{li2020infrared} & AAAI-20 & 49.92 & 89.79 & 95.96 & 50.73 & - & - & - & - & - & - & - & - & - & - & - & - \\
        cm-SSFT$^*$~\cite{lu2020cross} & CVPR-20 & 47.7 & - & - & 54.1 & 57.4 & - & - & \bf{59.1} & - & - & - & - & - & - & - & - \\
        MACE~\cite{ye2020cross} & TIP-20 & 51.64 & 87.25 & 94.44 & 50.11 & - & - & - & - & 57.35 & 93.02 & 97.47 & 64.79 & - & - & - & - \\
        DDAG~\cite{ye2020dynamic} & ECCV-20 & 54.75 & 90.39 & 95.81 & 53.02 & - & - & - & - & 61.02 & 94.06 & 98.41 & 67.98 & - & - & - & - \\
        HAT~\cite{ye2020visible} & TIFS-20 & 55.29 & 92.14 & 97.36 & 53.89 & - & - & - & - & 62.10 & 95.75 & 99.20 & 69.37 & - & - & - & - \\
        \hline
        CM-NAS & - & \bf{61.99} & \bf{92.87} & \bf{97.25} & \bf{60.02} & \bf{68.68} & \bf{94.92} & \bf{98.36} & 53.45 & \bf{67.01} & \bf{97.02} & \bf{99.32} & \bf{72.95} & \bf{76.48} & \bf{98.68} & \bf{99.91} & \bf{65.11} \\
        CM-NAS$^{\dagger}$ & - & 60.42 & 91.86 & 96.76 & 58.62 & 67.87 & 94.49 & 98.11 & 52.07 & 66.74 & 96.14 & 99.27 & 72.37 & 75.69 & 97.93 & 99.76 & 64.27 \\
        \myline
    \end{tabular}}
\end{table*}

\subsection{Comparisons with State-of-the-Art Methods}
We evaluate the proposed CM-NAS against current state-of-the-art VI-ReID methods on the SYSU-MM01 and the RegDB datasets.
The compared methods include Zero-Pad \cite{wu2017rgb}, TONE \cite{ye2018hierarchical}, HCML \cite{ye2018hierarchical}, cmGAN \cite{dai2018cross}, BDTR \cite{ye2019bi}, eBDTR \cite{ye2019bi}, HSME \cite{hao2019hsme}, D$^2$RL \cite{wang2019learning}, MSR \cite{feng2019learning}, AlignGAN \cite{wang2019rgb}, JSIA-ReID \cite{wang2020cross}, Xmodal \cite{li2020infrared}, MACE \cite{ye2020cross}, cm-SSFT \cite{lu2020cross}, DDAG \cite{ye2020dynamic} and HAT~\cite{ye2020visible}.
It should be noted that the multi-query setting of cm-SSFT uses multiple queries to constitute an auxiliary set to facilitate feature matching, which is infeasible in real scenarios and unfair to other methods.
Therefore, we only report the results of cm-SSFT under the single-query setting \cite{lu2020cross}.

\paragraph{Comparisons on SYSU-MM01.}
Table~\ref{table-2} displays the comparison results with state-of-the-art VI-ReID methods on the SYSU-MM01 dataset.
From the perspective of evaluation settings, \textit{i.e.} \emph{all-search}/\emph{indoor-search} and \emph{single-shot}/\emph{multi-shot}, we have two observations.
First, the results of all the methods under the \emph{indoor-search} setting are better than those under the \emph{all-search} setting. This is because the images in the \emph{indoor-search} setting only contain relatively brief indoor scenarios, while those in the \emph{all-search} setting have more complex in-the-wild scenarios.
Second, for a same method, the results under the \emph{multi-shot} setting are better than those under the \emph{single-shot} setting in the aspect of Rank-1 accuracies, while the phenomenon is opposite in the aspect of mAP scores.
This is due to the fact that each person identity contains 10 gallery images in the \emph{multi-shot} setting and only has 1 gallery image in the \emph{single-shot} setting.
As a result, for \emph{multi-shot}, it is prone to match one right sample according to the ranking of similarity scores, but it is difficult to match all right samples.
The following analyses are based on the \emph{single-shot}$\&$\emph{all-search} setting, since it is more challenging as mentioned in \cite{wang2019rgb,hao2019hsme}.

\begin{table}[t]
    \centering
    \caption{Comparisons with state-of-the-art methods on the RegDB dataset. CM-NAS$^{\dagger}$ means first searching a neural architecture on the SYSU-MM01 dataset and then training the searched neural network on the RegDB dataset.}
    \label{table-3}
    \resizebox{0.475\textwidth}{!} {
    \begin{tabular}{l|cccc|cccc}
        \myline
        \multirow{2}{*}{Method} & \multicolumn{4}{c|}{Visible to Infrared} & \multicolumn{4}{c}{Infrared to Visible} \\
        \cline{2-9}
        & R1 & R10 & R20 & mAP & R1 & R10 & R20 & mAP  \\
        \hline
        Zero-Pad~\cite{wu2017rgb}       & 17.74 & 34.21 & 44.35 & 18.90 & 16.63 & 34.68 & 44.25 & 17.82 \\
        HCML~\cite{ye2018hierarchical}  & 24.44 & 47.53 & 56.78 & 20.08 & 21.70 & 45.02 & 55.58 & 22.24 \\
        BDTR~\cite{ye2019bi}         & 33.56 & 58.61 & 67.43 & 32.76 & 32.92 & 58.46 & 68.43 & 31.96 \\
        eBDTR~\cite{ye2019bi}           & 34.62 & 58.96 & 68.72 & 33.46 & 34.21 & 58.74 & 68.64 & 32.49 \\
        HSME~\cite{hao2019hsme}         & 50.85 & 73.36 & 81.66 & 47.00 & 50.15 & 72.40 & 81.07 & 46.16 \\
        D$^2$RL~\cite{wang2019learning} & 43.4  & 66.1  & 76.3  & 44.1  & -     & -     & -     & -     \\
        MAC~\cite{ye2020cross}         & 36.43 & 62.36 & 71.63 & 37.03 & 36.20 & 61.68 & 70.99 & 36.63 \\
        MSR~\cite{feng2019learning}     & 48.43 & 70.32 & 79.95 & 48.67 & -     & -     & -     & -     \\
        AlignGAN~\cite{wang2019rgb}     & 57.9  & -     & -     & 53.6  & 56.3  & -     & -     & 53.4  \\
        JSIA-ReID~\cite{wang2020cross}  & 48.5  & -     & -     & 49.3  & 48.1  & -     & -     & 48.9  \\
        Xmodal~\cite{li2020infrared}    & 62.21 & 83.13 & 91.72 & 60.18 & -     & -     & -     & -     \\
        cm-SSFT$^*$~\cite{lu2020cross}      & 65.4  & -     & -     & 65.6  & 63.8  & -     & -     & 64.2  \\
        MACE~\cite{ye2020cross}         & 72.37 & 88.40 & 93.59 & 69.09 & 72.12 & 88.07 & 93.07 & 68.57 \\
        DDAG~\cite{ye2020dynamic}       & 69.34 & 86.19 & 91.49 & 63.46 & 68.06 & 85.15 & 90.31 & 61.80 \\
        HAT~\cite{ye2020visible}        & 71.83 & 87.16 & 92.16 & 67.56 & 70.02 & 86.45 & 91.61 & 66.30 \\
        \hline
        CM-NAS & \bf{84.54} & \bf{95.18} & \bf{97.85} & \bf{80.32} & \bf{82.57} & \bf{94.51} & \bf{97.37} & \bf{78.31} \\
        CM-NAS$^{\dagger}$ & 82.68 & 94.79 & 97.58 & 78.91 & 81.24 & 93.85 & 96.65 & 77.16 \\
        \myline
    \end{tabular}}
\end{table}

From the perspective of methods, our CM-NAS outperforms all the competitors by a large margin.
For instance, compared with the state-of-the-art HAT, we improve the Rank-1 accuracy and the mAP score by \textbf{6.70\%} and \textbf{6.13\%}, respectively.
Meanwhile, it is worth noting that the image size of the input of HAT is 288$\times$144, while that of our method is 256$\times$128.
The significant improvements with a smaller image size fully reveal the superiority of our method.
Furthermore, we also conduct a cross-dataset experiment to evaluate the generalization ability of the searched neural architecture.
We first search a neural architecture on RegDB, and then train the searched neural architecture on SYSU-MM01 with the loss function Eq.~(\ref{eq12}). Such a cross-dataset method is denoted as CM-NAS$^{\dagger}$ in Table~\ref{table-2}.
It is observed that CM-NAS$^{\dagger}$ still significantly surpasses all counterparts, unfolding the great generalization ability of the searched cross-modality architecture.

\paragraph{Comparisons on RegDB.}
As listed in in Table~\ref{table-3}, it can be seen that our CM-NAS has distinct advantages over state-of-the-art counterparts on RegDB.
Under the \emph{Visible to Infrared} setting, CM-NAS exceeds the state-of-the-art MACE by \textbf{12.17\%} and \textbf{11.23\%} in terms of the Rank-1 accuracy and the mAP score, respectively.
When switching to the \emph{Infrared to Visible} setting, compared with MACE, CM-NAS still improves the two indicators by \textbf{10.45\%} and \textbf{9.74\%}, respectively.
Besides, we also investigate the aforementioned cross-dataset method CM-NAS$^{\dagger}$, in which the neural architecture is first searched on SYSU-MM01 and then trained on RegDB.
Table~\ref{table-3} shows that CM-NAS$^{\dagger}$ is always superior to the compared methods, which further verifies the generalization ability of the searched architecture.

\section{Conclusion}
This paper has presented a novel CM-NAS to tackle the challenging VI-ReID.
We systematically investigate the manually designed neural architectures and find that appropriately separating BN layers can yield better performances.
This motivates us to develop a BN-oriented NAS algorithm that has the ability to automatically decide the separation of BN layers, searching the optimal architecture.
Extensive experiments on two popular datasets demonstrate the superiority of CM-NAS.
We expect this simple yet effective method will serve as a solid foundation to facilitate future research in VI-ReID.

\section*{Acknowledgment}
This work is supported by the National Key R\&D Program of China under Grant No. 2020AAA0103800.

{\small
\bibliographystyle{ieee_fullname}
\bibliography{egbib}
}

\end{document}